\documentclass[10pt,twocolumn,letterpaper]{article}
\usepackage{caption}
\usepackage[accsupp]{axessibility}  

\usepackage{cvpr}      

\definecolor{cvprblue}{rgb}{0.21,0.49,0.74}
\usepackage[pagebackref,breaklinks,colorlinks,allcolors=cvprblue]{hyperref}

\def\paperID{DriveX-12} 
\def\confName{CVPR}
\def\confYear{2025}

\title{\Large{Can Vision-Language Models Understand and Interpret Dynamic Gestures from Pedestrians? Pilot Datasets and Exploration Towards Instructive Nonverbal Commands for Cooperative Autonomous Vehicles}}

\author{
Tonko Bossen\\ 
University of California Merced\\
{\tt\small tbosse20@student.aau.dk}
\and
Andreas Møgelmose\\
Aalborg University\\
{\tt\small anmo@create.aau.dk}
\and
Ross Greer\\
University of California Merced\\
{\tt\small rossgreer@ucmerced.edu}
}

\begin{document}
\maketitle
\begin{abstract}
In autonomous driving, it is crucial to correctly interpret traffic gestures (TGs), such as those of an authority figure providing orders or instructions, or a pedestrian signaling the driver, to ensure a safe and pleasant traffic environment for all road users. This study investigates the capabilities of state-of-the-art vision-language models (VLMs) in zero-shot interpretation, focusing on their ability to caption and classify human gestures in traffic contexts. We create and publicly share two custom datasets with varying formal and informal TGs, such as `Stop', `Reverse', `Hail', etc. The datasets are ``Acted TG (ATG)" and ``Instructive TG In-The-Wild (ITGI)". They are annotated with natural language, describing the pedestrian's body position and gesture.
We evaluate models using three methods utilizing expert-generated captions as baseline and control: (1) caption similarity, (2) gesture classification, and (3) pose sequence reconstruction similarity. Results show that current VLMs struggle with gesture understanding: sentence similarity averages below 0.59, and classification F1 scores reach only 0.14–0.39, well below the expert baseline of 0.70. While pose reconstruction shows potential, it requires more data and refined metrics to be reliable.
Our findings reveal that although some SOTA VLMs can interpret zero-shot human traffic gestures, none are accurate and robust enough to be trustworthy, emphasizing the need for further research in this domain. We make our code publicly available at \hyperlink{https://github.com/tbosse20/gest_VLM_eval}{github.com/tbosse20/gest\_VLM\_eval}
\end{abstract}
\section{Introduction}
\label{sec:intro}
Scene understanding and decision-making in autonomous driving rely on the ability of systems to predict the future location of moving objects \cite{salzmann2020trajectron++, prakash2021multi, greer2021trajectory}. Still, a limitation of safe autonomy lies in understanding the gestures of surrounding humans. This decreases the safety and trust of these systems in interactive traffic scenarios.

While physical constraints of motion may inform trajectory prediction methods for dynamic objects, in this research, we approach the challenge of intent prediction \cite{alofi2024pedestrian}, where motion must be anticipated before it begins. However, our research considers not only the intention of an individual agent but also the intentions the agent imposes on others in the form of instructions \cite{roy2024doscenes}. We make a clarifying distinction in uses of the word `intention': borrowing from attention-based learning architecture terminology, an agent's self-intention describes their intended future actions, while an agent's cross-intention describes the future actions of other agents as intended by the observed agent. 

\begin{figure}
    \centering
    \includegraphics[width=0.4\textwidth]{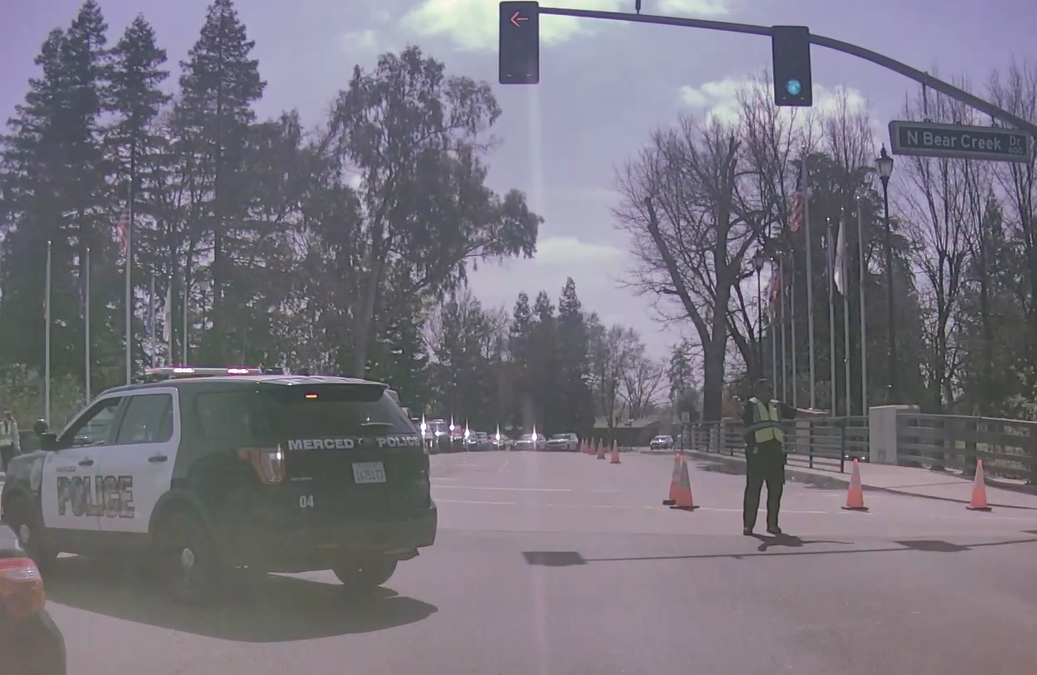}
    \caption{In autonomous driving scenarios, navigation instructions may come from pedestrians' dynamic, nonverbal gestures. Interpreting and responding to such gestures is vital for safe autonomous driving.}
    \label{fig:realsituation}
\end{figure}

An agent's intentional gestures may communicate intent, querying, and instruction, and often require comprehensive scene understanding to ensure a safe response, especially when other independent agents are present in the scene, making the complete scene motion less predictable. In ideal settings, a scene may have a formalized focal point, such as a law enforcement or traffic-directing officer, such as in Fig. \ref{fig:realsituation}, whose authority reduces the complexity of interpreting the scene and selecting gestures to follow. Understanding traffic gestures is crucial for autonomous vehicles (AVs) to function in these \textit{formally-directed} settings, where the ability to follow instructions is essential. In non-authoritative or \textit{informal} situations, though there is no direct jurisdiction of control. Pedestrians can signal their own intent, expectancy of the ego drivers' intent, or information about the scene. For example, acting as a vantage point for occluded areas. While explicit communication from pedestrians occurs in only 2.7\% of road-crossing events \cite{dey2017pedestrian}, it can enhance the experience and safety of pedestrians and drivers by increasing the scene understanding. In this research, we consider that the pedestrian is not just a passive agent in the traffic scene, but a source of deliberate, intentional information and instructions to the ego vehicle, whether or not they are ordained with authority. 

Human gestures can be subtle, yet through common sense and experience, people can often understand one another using only gestures. However, the interpretation of gestures can vary between countries and cultures, adding to the complexity of intention understanding \cite{rasouli2019autonomous}. The wide range of subtle variations in gestures may be difficult to capture in any limited-size training dataset, and further may be difficult to extrapolate in meaningful ways through pose estimation subtask modules towards gesture understanding. So, in this study, we seek to understand if vision-language models (VLMs) trained on foundation-model-scale data may encode this information to be recovered in a zero-shot manner. Teaching AV systems to interpret pedestrians' incoming gestures can create a more maneuverable, efficient, and safer road environment. This research explores how AI can perceive and utilize pedestrian gestures to make better decisions.



\begin{figure}
    \centering
    \begin{subfigure}[t]{0.2\textwidth}
        \centering
        \includegraphics[width=0.734\textwidth]{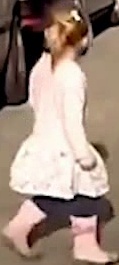}
        
        \caption{Girl pedestrian walking in front of the ego car, looking down the road, with her arms down her side (crop).}
        \label{fig:girl}
        
    \end{subfigure}
    \hspace{0.03\textwidth}
    \begin{subfigure}[t]{0.2\textwidth}
        \centering
        \includegraphics[width=\textwidth]{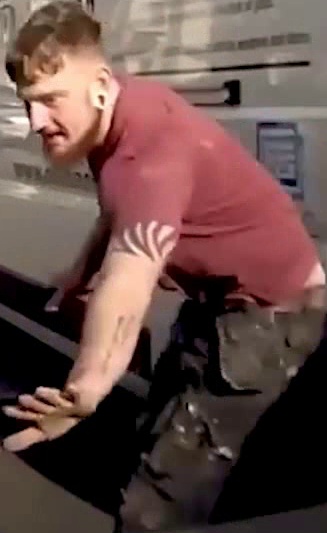}
        
        \caption{Man pedestrian gesture ego driver to `Stop' with one arm (crop).}
        
        \label{fig:man}
    \end{subfigure}
    
    \caption{Pedestrians from frame 71 from video 0153 of the COOOL dataset \cite{alshami2024coool}. Zero-shot analysis of these pedestrians using VLMs fails to capture the significance of their gestures (or non-gestures) toward the traffic scene.}
    
    \label{fig:coool_frames}
\end{figure}

We perform an initial evaluation using a few online demos to gain insight into this task. Figures \ref{fig:girl} and \ref{fig:man} show two pedestrians extracted from left-to-right from a real traffic scene of the COOOL dataset \cite{alshami2024coool}. Fig. \ref{fig:man} provides a preliminary example that hints toward the general incapability of VLMs to caption traffic gestures. The VLM received the image along with the prompt: \textit{``What is this pedestrian gesturing?"} using available online VLM demos 
\textit{BLIP2}\footnote{\hyperlink{https://huggingface.co/spaces/hysts/BLIP2}{huggingface.co/spaces/hysts/BLIP2}},
\textit{VideoLLaMA2}\footnote{\hyperlink{https://huggingface.co/spaces/lixin4ever/VideoLLaMA2}{huggingface.co/spaces/lixin4ever/VideoLLaMA2}},
\textit{VideoLLaMA3}\footnote{\hyperlink{https://huggingface.co/spaces/lixin4ever/VideoLLaMA3}{huggingface.co/spaces/lixin4ever/VideoLLaMA3}},
\textit{VideoLLaMA3-Image}\footnote{\hyperlink{https://huggingface.co/spaces/lixin4ever/VideoLLaMA3-Image}{huggingface.co/spaces/lixin4ever/VideoLLaMA3-Image}}, and
\textit{ChatGPT-4o}\footnote{\hyperlink{https://chatgpt.com/}{chatgpt.com}}
.

The output prompting these images varied from \textit{``The person in the video is gesturing something that resembles a farting sound or action."} from VideoLLaMA2, to \textit{``...their body posture suggests they might be trying to stop something or someone, maintain balance ..."} from ChatGPT-4o. However, even the slightly more promising ChatGPT model was quick to hallucinate or misinterpret gestures when presented with the same prompt accompanying Fig. \ref{fig:girl}. This image was a crop of a little girl walking in front of the ego driver, facing away from the ego driver with her arms down her sides. Still, ChatGPT's output stated she was raising one arm and waving or pointing. This output expressed further insecurity due to image quality despite the visual clarity. These brief examples highlight the impetus for a more thorough research into the limitations of these models in recognizing gesture-based human communication.

\section{Related Work}
\label{sec:related}
Existing VLMs can generate various output forms supporting autonomous driving, ranging from natural language captions that might inform trajectory generation \cite{atakishiyev2023explaining, keskar2025evaluating}, selection of specific control commands \cite{shao2024lmdrive}, novelty detection \cite{greer2024towards, greer2025language}, or even end-to-end learning of direct waypoint trajectories \cite{xu2024drivegpt4, chen2024end}. A particular class of models that are most relevant for gesture recognition are Video Foundation Models (ViFMs) due to the temporal nature of gesture-based communication \cite{madan2024foundation}.In this section, we elaborate on some existing VLM techniques and datasets toward the safe navigation task.


CoVLA combines VLMs and object detection to generate a caption of the scene, which is used to learn a trajectory projected upon the scene image \cite{CaVLA}. The usage of VLMs is considered for the end-to-end training of networks, which may naturally include gesturing pedestrians. Still, we suggest that scene agent gestures can be so fundamental to control decisions that implicit end-to-end learning may not be sufficient. Further, the existence of confounding factors (e.g., a green light with a traffic controller simultaneously indicating forward motion) may inhibit the model from learning causality between scene object patterns and control decisions \cite{greer2023robust, greer2022salience, greer2023salient}. Further, models like CoVLA and DriveLLaVA \cite{zhao2024drivellava} learn precise trajectory outputs rather than abstract direction commands; the alignment of the latter may be more suited to learning from gesture, though it is our intention in developing methods for such gesture feature extraction that this may also be used modularly within trajectory learning pipelines.


While existing RGB-media datasets around pose and gesture focus on `action', `pose estimation', `sign language', and `hand gesture' \cite{amir2017low, escalera2013multi, multi-modal-gesture-recognition, ruffieux2014survey}, there is a gap in data within the navigation/traffic gestures domain, which predominantly utilize the upper-body. Non-domain datasets still provide utility for general pose classification using VLMs, expanding the general understanding of the capabilities of these models. Within the domain, the Traffic Control Gesture (TCG) dataset includes 250 sequences of 3D body skeleton \cite{wiederertraffic}. While this dataset is highly pertinent to our study, it solely provides 3D pose annotations without the accompanying visual data, limiting its applicability in VLMs. While the anomaly detection dataset COOOL \cite{alshami2024coool}  contains a few scenes of pedestrian gesture-based communication to the ego driver, the scenes are unannotated and are too few in number for a gesture-specific analysis.

A promising advancement in zero-shot recognition of hand gestures using image data is GestLLM \cite{kobzarev2025gestllm}, which integrates large-language models with pose-based feature extraction. GestLLM system showed robust performance in hand gesture recognition, providing one path towards improving the zero-shot VLM issues studied in this paper.
\section{Methodology}
This study seeks to understand the zero-shot capabilities of VLMs in recognizing and responding to static, dynamic, and composite human traffic gestures in RGB videos with physical body descriptions and contextual interpretations within driving scenarios. To do this, we utilize three evaluation methods \textit{1. Embedded Similarity}, \textit{2. Classification}, and \textit{3. Reconstruction} to quantify the performance of models in converting human gestures and motion to text and derivative meaning.

By using VLMs, \textit{intentional} gestures (e.g., hand gestures) are emphasized with \textit{accompanying} (e.g., body language and facial cues) second-hand information, which may or may not agree with the manual (i.e., hand-communicated) intention.

\subsection{Dataset}\label{sec:dataset}
We create and publish two datasets for this study\footnote{Link to datasets in README: \hyperlink{https://github.com/tbosse20/gest_VLM_eval}{github.com/tbosse20/gest\_VLM\_eval}}:
1. `Acted Traffic Gestures' (ATG) with a single actor portraying various gestures to the camera as a hypothetical ego vehicle, used for this evaluation. 2. `Instructive Traffic Gestures In-the-Wild' (ITGI) is a real-world encounter of an ego vehicle with traffic conductors, filmed from four synchronized dash cameras for a multi-directional surround view, added as additional data.

{\bfseries\fontsize{10pt}{12pt}\selectfont Acted Traffic Gestures (ATG)} The ATG dataset features a single actor gesturing towards a static camera recorded at 30 FPS. The camera is 1.6 meters above the ground and 1 - 2 meters from the participant, acting as a vehicle dash-cam. It is recorded inside a closed room against a white wall, to lock parameters and reduce noise. The dataset includes 8 short videos with gestures for `Idle', `Reverse', `Go', `Stop, pass', `Follow', `Forward', `Stop, go', and `Hail', ranging from 1 to 4 seconds. The `Stop, pass' and `Stop, go' are composed gestures of `Stop' to the ego driver, and `Pass' or `Drive' to other vehicles. These gestures are acted out as an unofficial but naturalistic traffic guide rather than following any municipality's official traffic warden gestures. \textit{In this report, we detail the initial dataset properties at the time of writing, and the dataset continues to be extended with varying environments, distances, and gestures to enable research across a wider range of scenes and parameters. Updated dataset details are available in the repository README.}

\begin{figure}
    \centering
    \includegraphics[width=0.2\textwidth]{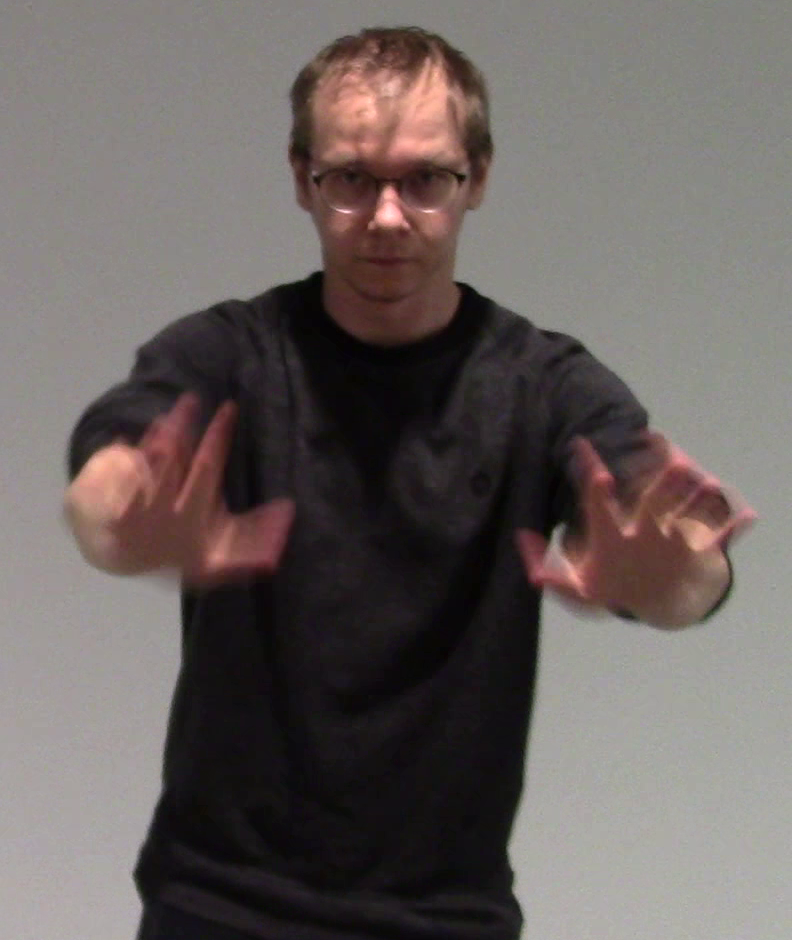}
    \caption{Frame 18 from the ``Reverse" command gesture video. This frame's set was annotated with the caption, \textit{"The pedestrian is standing in front of the ego driver. They are facing their torso and head towards the ego driver. They are moving their flat palms back and forth towards the ego driver, gesturing for it to reverse. They are moving slightly to the side." }}
    \label{fig:reverse_frame}
\end{figure}

Ground truth annotations were made by a licensed driver with oracle knowledge of the underlying gestures instructed to the actor. This annotator reviewed each video at 8-frame intervals, without overlapping segments. The traffic gesture label is described from both the pedestrian's and the driver's perspectives combined in each annotation, and it is interpreted in terms of the pedestrian's intended communication towards the ego driver or other drivers. Additionally, \textit{expert-generated} captions were made by additional licensed drivers. They serve both as a `baseline' to assess the overall effectiveness and accuracy of the evaluation method, and as `supplementary' ground truths to emphasize the intended meaning of the gestures, rather than the specific wording. The `instructions' considered when generating the ground truth is formulated as follows: \textit{"Describe the pedestrians' body posture focusing on their arm position and movement relative to both themselves (e.g., at their side, in front of them) and the ego driver (e.g., towards the ego driver, left of the ego driver), their hand position and shape (e.g., flat hand faced downward), and the orientation of their body and face (e.g., facing to the left). Include an interpretation of potential gestures and their intended recipient (e.g., signaling to stop, requesting to pull over)."} A ground truth caption example and a corresponding frame from an 8-frame sequence are provided in Fig. \ref{fig:reverse_frame}. In addition to the short-term motion analysis annotations, we provide a complete caption describing the gesture for each complete video.

{\bfseries\fontsize{10pt}{12pt}\selectfont Instructive Traffic Gestures In-the-Wild (ITGI)} The dataset is collected while driving around town during a bike race, which involved police enforcement guiding in intersections. It varies from minor intersections with a single idle police officer and a few cones, to light intersections blocked by police vehicles with multiple officers guiding cars. The scenes are all set in formal settings, including official traffic regulation gestures and more casual/unofficial gestures. It was recorded from a Tesla with four built-in cameras: front, back, right-back, and left-back at 36 FPS. The data consists of 18 videos ranging from 8 seconds to 2 minutes. We provide this dataset without annotation.


\subsection{Models and Setting}
The selected models evaluated in this study are \textit{VideoLLaMA2} \cite{damonlpsg2024videollama2}, \textit{VideoLLaMA3} \cite{damonlpsg2025videollama3}, and \textit{Qwen2} \cite{Qwen2-VL}. The VideoLLaMA models were selected for their recent notable performance in the CoVLA paper \cite{CaVLA}, and Qwen's near relationship as a subset of VideoLLaMA. We limited our evaluation over models which are available to run locally on an edge device without internet access or API charge, which excludes candidate model ChatGPT-4o. All evaluations were run on one NVIDIA® GeForce RTX™ 4090 24GB compatible with \texttt{float16} precision and Flash-Attention 2.0 \cite{dao2023flashattention2fasterattentionbetter}. We use a model temperature of 0.2 and 512 maximum new tokens. 

We use 8-frame samples per caption. This window span was selected as a baseline for experiments to engage frame-to-frame captioning with temporal context. The frame rate was chosen as an estimate of the necessary temporal information to reasonably represent a gesture, an area we highlight for future research. The correct frame rate is a study in itself, as traffic gestures in this dataset can vary from 0.1 to 5 seconds long, and multiple gestures can be combined into one command interpretation, making it challenging to capture a complete gesture and its communicative intent within a short window of eight frames.

\subsection{Prompting}
For the purpose of this research, we craft a series of prompts intended to extract varying types of information from the VLM. We acknowledge that partial information may enable chain-of-thought reasoning as opposed to zero-shot task success, and that analysis of subtasks may be valuable toward ongoing research. Two central pieces of information included in the prompts are the `Context' (a driving autonomous vehicle from a dash-cam perspective), and the `Objective' (retrieving information for safe, intention-aligned decision-making). Additionally, to the point of subtask analysis, our prompts seek to extract either an `Explanation' of the agent's physical motion involved in the gesture, for an LLM to interpret downstream, or a direct interpretation of the gesture from the VLM itself. This is useful in cases when the VLM does not grasp the concept of the meaning of the gestures, but could explain the agent's movements well enough for an LLM to interpret.

To give the models a broader chance to successfully caption the gesture correctly, varying text prompts were used to evaluate each model. The five varying prompts used are referred to as `Blank' (\textit{The prompt is left empty; the images alone serve as prompt}), `Determine' (\textit{"Determine what gesture the pedestrian is making."}), `Body' (\textit{"Provide a detailed explanation of the pedestrian's body posture and movements."}), `Context' (\textit{"You are an autonomous vehicle navigating a road. Determine what gesture the pedestrian is making."}), and `Objective' (\textit{"You are an autonomous vehicle navigating a road. Determine what gesture the pedestrian is making. Your response will be used by an AI system to make real-time driving decisions."}), which vary in understanding and focus. While ideally the accurate interpretation of motion should be enough to inform the meaning of the gesture, in some prompts we suggest that the VLM consider the context of the 3D driving scene.
\section{Evaluation}
The evaluation section consists of the three evaluation methods: `Embedded Similarity', `Classification', and `Reconstruction'. We approach this topic from multiple angles, in different degrees of abstractions and focus points. The evaluations are conducted using the ATG dataset. Each section provides a description of the method and the results. Discussions are combined in Section \ref{sec:discussion}.

\subsection{Embedded Similarity} \label{eval:sim}
\begin{figure}
    \centering
    
    \makebox[\linewidth][c]{
        \includegraphics[width=1.05\linewidth, trim=0.25cm 0.25cm 0.25cm 0.25cm, clip]{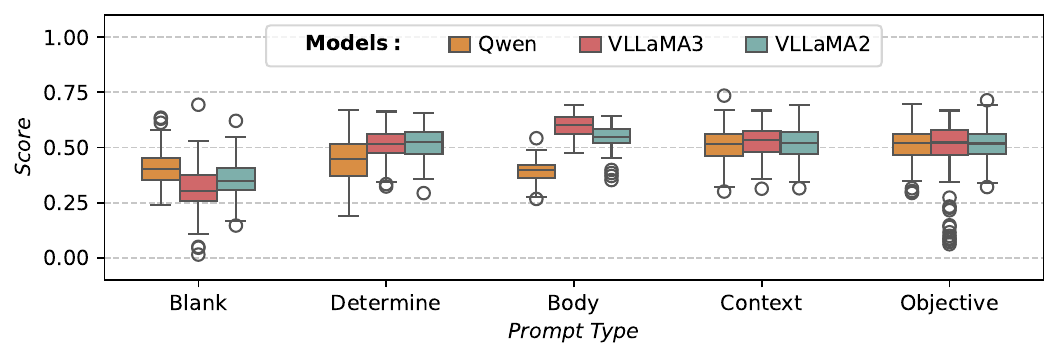}
    }
    
    \caption{Cosine similarity between \textit{expert} and \textit{ground truth} captions, across prompts. The varying prompts do not score higher than 0.75 in any samples, with means from 0.31 in VideoLLaMA3 `Blank' to 0.59 in VideoLLaMA3 `Body'. This indicates that specific prompts output more accurate captions than non-VLM captions. `Context' and `Objective' are around 0.51 across all models, while `Body' is only more accurate when used with the VideoLLaMA models.}
    
    \label{fig:prompt_types}

\end{figure}

\begin{figure}
    \centering 

    \makebox[\linewidth][c]{
        \includegraphics[width=1.05\linewidth, trim=0.25cm 0.25cm 0.25cm 0.25cm, clip]{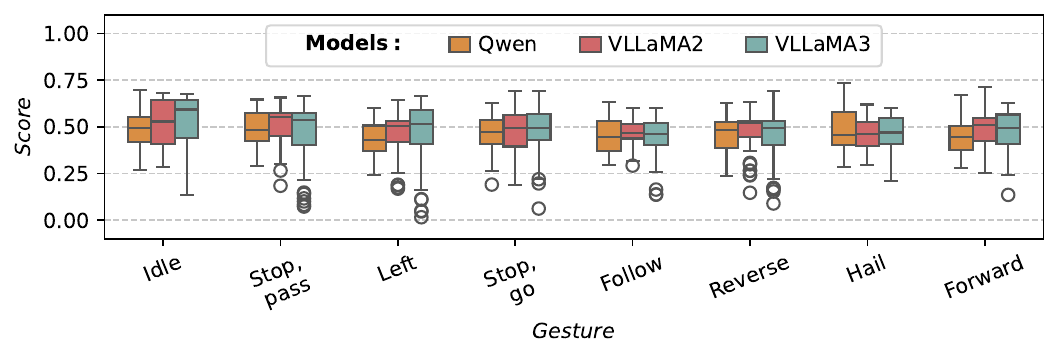}
    }
    
    \caption{Cosine similarity between \textit{expert} and \textit{ground truth} captions, across gestures. No gesture interpreted by a VLM reaches a score of 0.75 in any sample. The means vary from 0.43 in Qwen `Left' to 0.52 in VideoLLaMA3 `Idle'. Further, no specific gesture stands out as most readily interpretable.}
    
    \label{fig:gestures}
    
\end{figure}

This evaluation method seeks to understand the similarity between the generated caption and the ground truth caption. The generated captions' similarity was evaluated by embedding them using the SBERT encoder (\texttt{all-MiniLM-L6-v2}) \cite{reimers2019sentence} and measuring their similarity to the ground truth embeddings using \textit{Cosine Similarity} \cite{salton1983introduction}. The cosine similarity scores of the expert-generated captions serve as a baseline for assessing the overall effectiveness and accuracy of the evaluation method. To ensure that the results are not biased by a particular gesture or prompt, the outcomes are analyzed separately by prompt type in Fig. \ref{fig:prompt_types} and by gesture in Fig. \ref{fig:gestures}. 

\begin{figure}
    \centering
    \makebox[\linewidth][c]{
        \includegraphics[width=1.05\linewidth, trim=0.25cm 0.3cm 0.25cm 0.25cm, clip]{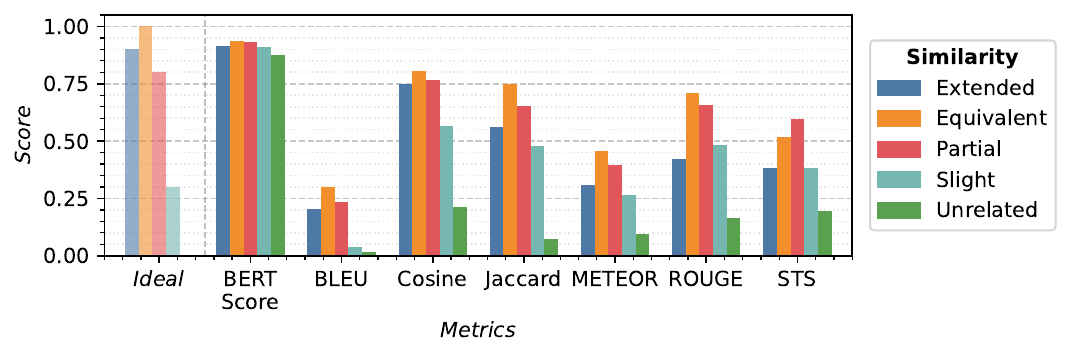}
    }
    
    \caption{We validate the selected encoder and metric by testing their ability to illustrate the expected similarity trend. The `Ideal' ratio trend is illustrated on the left, used to validate the most suitable metric. This involves comparing a target sentence to a series of rephrased versions with progressively decreasing similarity levels. Two target captions were formulated about the same hypothetical scenario (e.g., ``\textit{A person signals the ego driver to stop, by putting their hand towards the ego driver.}"). Each similarity level contains two rephrases cross-validated against both target captions to reduce sentence noise. The rephrases span five levels of similarity: `Extended' with additional information, which can be difficult to know it is irrelevant (noise) or incorrect (false positive) information (e.g., ``\textit{A pedestrian raises their hand towards the ego driver to stop traffic. They are looking scared and in need of help.}"), `Equivalent' with the same information (e.g., ``\textit{A pedestrian raises their hand towards the ego driver to stop traffic.}"), `Partial' with partially equivalent information (e.g., ``\textit{A person raises their hand towards the ego driver.}"), `Slight' which is missing important details (e.g., ``\textit{A human gestures to the ego driver."}), and `Unrelated' information (e.g., ``\textit{The sky is blue and the sun is shining.}").}
    
    \label{fig:metric_valid}
    
\end{figure}

To argue for the selected metric, we validated multiple metrics on decreasing similarity rephrasing of a target caption. We expect to see a trend of metric values decreasing as caption quality decreases (relative to the original target caption). We illustrate the results of this experiment in Fig. \ref{fig:metric_valid}, with `Ideal' as the optimal metric trend, which is hardcoded. The sampled metrics are BERT score \cite{zhang2019bertscore} (not to confuse with the SBERT encoder), BLEU \cite{papineni2002bleu}, Cosine Similarity \cite{salton1983introduction}, Semantic Textual Similarity (STS) \cite{reimers2019sentence}, Jaccard \cite{jaccard1901etude}, METEOR \cite{banerjee2005meteor}, and ROUGE \cite{lin2004rouge}.

SBERT was selected as the implemented encoder, due to its 74\% difference in cosine similarity between `Equivalent' and `Unrelated' validation captions, in contrast to Vanilla BERT \cite{wolf-2020-transformers}, showing only an 11\% difference. This makes it easier to distinguish. Other evaluation metrics remained consistent regardless of the encoder used. 

We do not expect a similarity score of 1.0, as the wording has to be exact for this to happen. That is neither likely nor intended. The similarity score is considered highly accurate at around 0.80, the score of the `Equivalent' validation captions, by looking at the validation at Fig \ref{fig:metric_valid}. Scores near 0.75 are considered moderately accurate as `Extended' or `Partial'. Scores near and below 0.55 are considered low accuracy as `Slight' similarity.

\begin{figure}
    \centering
    \begin{subfigure}[t]{0.48\linewidth}
        \centering
        \includegraphics[height=4.35cm, trim={5 0 0 5}, clip]{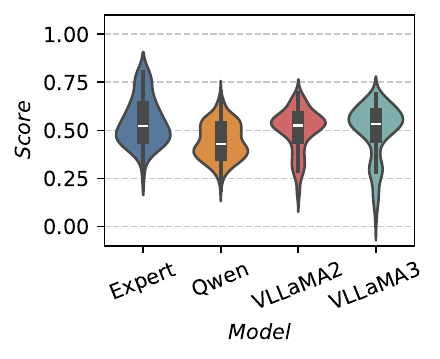}
        \caption{\textit{Expert} captions as baseline}
        \label{fig:metrics}
    \end{subfigure}
    \hfill
    \begin{subfigure}[t]{0.48\linewidth}
        \raggedleft
        \includegraphics[height=4.35cm, trim={50 0 5 5}, clip]{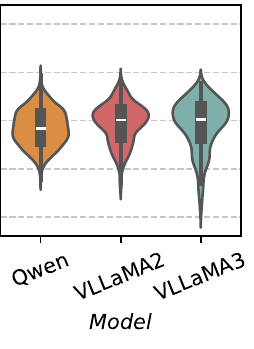}
        \caption{Performance averaged across all human annotations}
        \label{fig:metrics_to_humanngt}
    \end{subfigure}
    
    \caption{Cosine similarity with the generated captions relative to only the ground truth captions (\ref{fig:metrics}, left) and both the ground truth and expert captions (\ref{fig:metrics_to_humanngt}, right). They are visualized together to illustrate the decreasing standard deviation due to the denoising. 
    In \ref{fig:metrics} VideoLLaMA2 and VideoLLaMA3 show a slightly higher Q1 than the expert-generated captions. However, experts' mean is 0.54, while the mean of VideoLLaMA2 is 0.50, and VideoLLaMA3 is 0.49. The mean of Qwen is 0.44.
    In \ref{fig:metrics_to_humanngt} Qwen is increased to 0.46, but VideoLLaMA2 and VideoLLaMA3 are decreased to 0.48 and 0.47. This again shows a diversity in the non-VLM captions, and there are multiple ways to describe a certain gesture.}
    
    \label{fig:metrics_combined}
    
\end{figure}

\begin{table}
    \centering
    \renewcommand{\arraystretch}{1.0}
    \resizebox{0.5\textwidth}{!}{
    \begin{tabular}{|c|p{0.75\linewidth}|}
        \hline
        \textbf{Model} & \textbf{Caption} \\ \hline
    
        \begin{tabular}[t]{@{}l@{}}\footnotesize Ground \\[-0.3em] \footnotesize Truth\end{tabular} & \footnotesize\textit{``The pedestrian is standing in front of the ego driver. They are facing their torso and head towards the ego driver. They are moving their flat palms back and forth towards the ego driver, gesturing it to reverse. They are moving slightly to the side."} \\\hline
        
        \footnotesize Expert & \footnotesize\textit{``The person pushes their hands that are facing the camera outwards and brings them back in almost as if their signaling to stop or slow down"} \\\hline
        
        \begin{tabular}[t]{@{}l@{}}\footnotesize Qwen\end{tabular} & \footnotesize (abr.)~\footnotesize\textit{``The person in the image appears to be making a stop or "no" gesture with both hands extended forward and fingers spread apart, palm facing outward ..."} \\\hline
        
        \begin{tabular}[t]{@{}l@{}}\footnotesize VLLaMA2\end{tabular} &  \footnotesize\textit{``The pedestrian is making a stop gesture."} \\\hline
        
        \begin{tabular}[t]{@{}l@{}}\footnotesize VLLaMA3\end{tabular} & \footnotesize\textit{``The pedestrian is making a gesture with his hands."} \\\hline
    
    \end{tabular} }
    
    \caption{Examples of captions from the video sequence ``Reverse" from frame 16 - 24 given the ``Determine" prompt. We manually analyze the captions focusing on body movement, direction description, biases, and common traits. We see that the outputs from VideoLLaMA are short, with only VideoLLaMA2 giving a gesture response, albeit inaccurate. The example from Qwen is only a snippet, since its response is quite long, and it also responds inaccurately with `Stop'. Examples of generated captions from whole videos are shown in Table \ref{tab:determinereverse_video}.}
    
    \label{tab:determinereverse}
\end{table}

\begin{table}
    \centering
    \renewcommand{\arraystretch}{1.0}
    \resizebox{0.5\textwidth}{!}{
    \begin{tabular}{|c|p{0.75\linewidth}|}
        \hline
        \textbf{Model} & \textbf{Caption} \\ \hline
    
        \begin{tabular}[t]{@{}l@{}}\footnotesize Ground \\[-0.3em] \footnotesize Truth \end{tabular} & \scriptsize (abr.)~\footnotesize\textit{``..They look up, and move their hands back and forth towards me at their chest, indicating me to reverse.."} \\\hline
        
        \begin{tabular}[t]{@{}l@{}}\footnotesize Qwen \end{tabular} & \footnotesize (abr.)~\footnotesize\textit{``The pedestrian appears to be waving or gesturing with both hands as they walk past the camera. The movement suggests that they might be saying hello, goodbye, or simply acknowledging someone.."} \\\hline
        
        \begin{tabular}[t]{@{}l@{}}\footnotesize VLLaMA2\end{tabular} & \footnotesize\textit{``The pedestrian is making a stop gesture."} \\\hline
        
        \begin{tabular}[t]{@{}l@{}}\footnotesize VLLaMA3\end{tabular} & \footnotesize\textit{``The pedestrian is making a gesture with his hands."} \\\hline
    
    \end{tabular} }
    
    \caption{Examples of captions from the whole `Reverse' video parsed the `Determine' prompt (without expert-generated caption). The VideoLLaMA models output the same responses even when given more video information. Qwen outputs another inaccurate gesture interpretation. Examples of captions per 8 frames are shown in Table \ref{tab:determinereverse}.}
    
    \label{tab:determinereverse_video}
\end{table}

\subsection{Classification}\label{eval:cls}
Arguing that the semantic sentence comparison is complex to generate and evaluate, we reduce the model task to predicting a precise answer with complete interpretation. The possible classes are the same as the videos provided, plus additional common traffic gestures, making a total of 9 classes: 
`Follow',
`Hail',
`Forward',
`Right',
`Left',
`Idle',
`Reverse',
`Stop', and
`Other'. The provided videos include more complex gestures like `Stop, pass', but to simplify it, this evaluation method only focuses on the gesture towards the ego driver. The prompt was formulated with the context, steps to identify the matter, output format, and possible classes, each with a short description.

\begin{table}
    \centering
    \setlength{\tabcolsep}{4pt}
    \renewcommand{\arraystretch}{1.2}
    \resizebox{0.45\textwidth}{!}{
    \begin{tabular}{|l|c|c|c|c|c|c|c|c|c|} \hline

\textbf{Model} &
\textbf{Accuracy}  $\uparrow$ &
\textbf{Precision} $\uparrow$ &
\textbf{Recall}    $\uparrow$ &
\textbf{F1-Score}  $\uparrow$
\\ \hline

\textbf{Expert}  & \textbf{0.72} & \textbf{0.71} & \textbf{0.72} & \textbf{0.70} \\ \hline
\textbf{Qwen}    & 0.33 & 0.11 & 0.33 & 0.17 \\ \hline
\textbf{Vllama2} & 0.15 & 0.28 & 0.15 & 0.14 \\ \hline
\textbf{Vllama3} & 0.52 & 0.32 & 0.52 & 0.39 \\ \hline

    \end{tabular} }
    \caption{Classification accuracy with 9 classes on ATG with 8-frame interval. All VLMs have difficulty interpreting gestures correctly, even toward this reduced task, with a best F1 score lower than 0.40. This shows they can perceive and interpret traffic gestures to some extent, but are unreliable toward autonomous driving in their current form. Processed expert annotations successfully classify `Left', `Reverse', and `Stop' with F1 above 0.80. Qwen captions predict `Stop' 56 times and `Hail' 6 times out of 62. VideoLLaMA2 predicts `Hail' 53 times out of 62. VideoLLaMA3 confuses `Left' with `Forward' and `Reverse' with `Stop', and has `Hail' and `Stop' with an F1 above 0.70.}
    \label{tab:arocss_gestures}
\end{table}

\subsection{Reconstruction} \label{eval:reconstruction}
\begin{figure}
    \centering
    \includegraphics[width=0.25\textwidth]{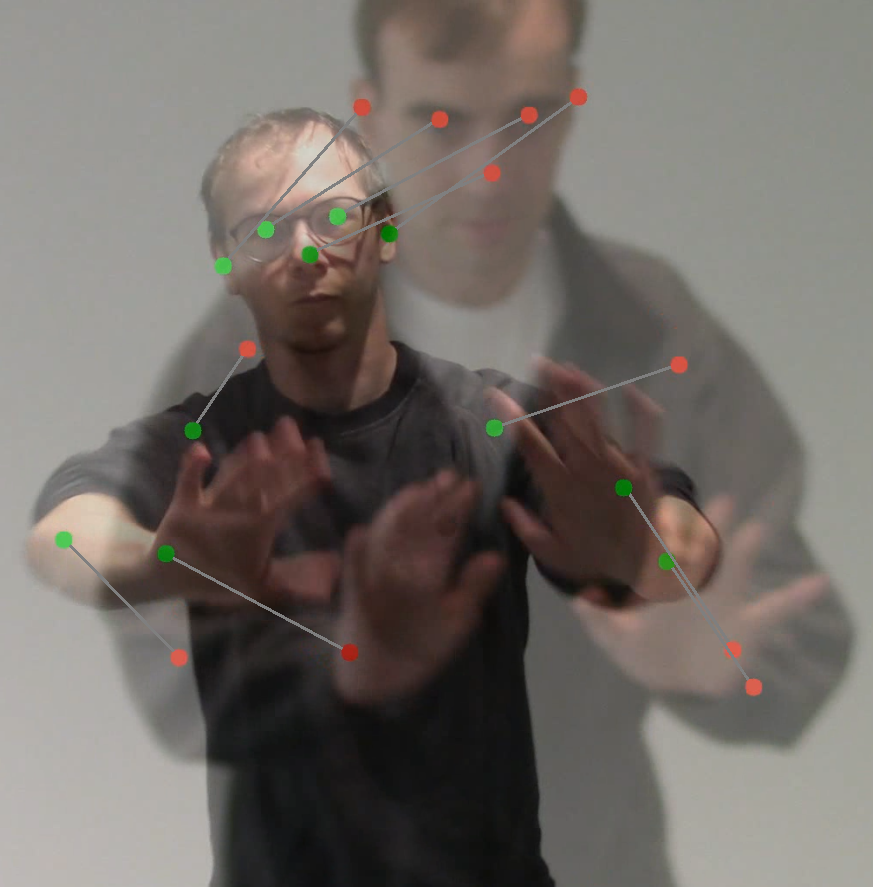}
    \caption{Reconstructed gesture using the ground truth caption, overlay upon the original `Reverse' video \textit{(timestamp 00:03)}.}
    \label{fig:merged}
\end{figure}

We design one additional method of evaluation of machine-generated captions, built around the premise that precise and detailed language should carry sufficient information for a movement to be reenacted accurately. For example, ``raise one arm" can have many meanings, while "raise your left arm right above your head" is much more precise and less likely to be misunderstood. An example is visualized in Fig. \ref{fig:merged}.

\begin{figure}
    \centering
    \includegraphics[width=0.45\textwidth]{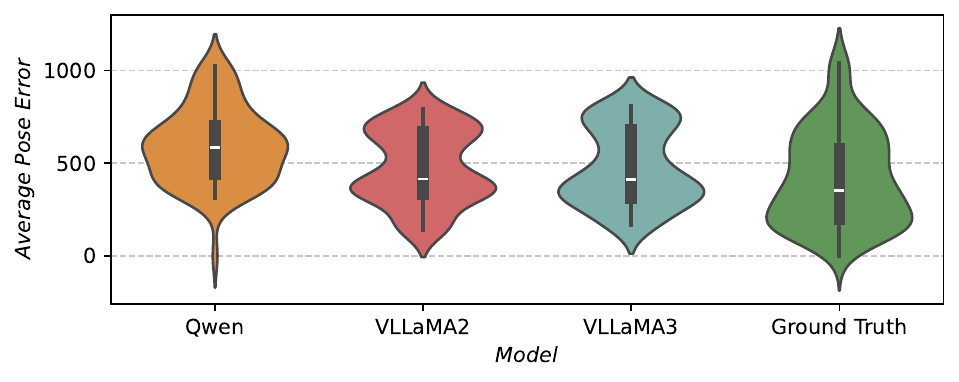}
    \caption{The results show MSE for every frame with the caption from the whole video `Reverse'. The participant is not the same height (Fig. \ref{fig:merged}) as in the original video, which makes it more challenging to be accurate, but leaves an opportunity for future research as an evaluation method. 
    As a further complication for future consideration, it was found that it was either not temporally correct to reenact the gesture as it was read out loud, or could miss minor details if reenacted after reading the whole caption. However, the ground truth scored better than the VLM caption reenactments. The Qwen model had a mean score around 600, VideoLLaMA models around 400, and ground truth around 300. A few frames scored around 0, an exact pose in the same frame. However, considering the difficulty of complete synchronous pose matching, we consider this largely a coincidence with the large number of frames evaluated.}
    \label{fig:reconstruct_results}
\end{figure}

This evaluation was constructed by reading the captions aloud to a participant, who would reenact the described movements as informed in the caption with their interpretation. This reenacted scene was compared with the original video to compute a coarse evaluation metric, using pose estimation and MSE upon the equivalent pose points. The reconstructed videos were cut only to contain the movement and sped up to match the length of the original video. Results of this analysis are shown in Fig. \ref{fig:reconstruct_results}.

\section{Discussion} \label{sec:discussion}
Due to the desired application of the VLMs, we explored prompting for detailed captions and closed-set interpretation classes. Some of these models vary considerably in their capabilities of captioning human gestures. Occasionally, some captions would be spot-on, or more accurate to certain gestures like `Stop' and `Hail', but in all three evaluation methods, the VLMs had a lower score than the expert baseline. As the expert baseline was also considerably low, ambiguity and evaluation strategies merit future research for reliable interpretation of human traffic gestures.

For data collection and experimental design, we note that the expert-generated captions are biased in knowing the whole video and the name of the video, which gives more information about the gesture than the individual sequences. To increase fidelity, each expert should too have varying prompts like the VLMs and the ground truth captions, but this would require more participants to avoid `preknowledge'-biases. This further begs the additional question of bias: will pedestrians use the same gestures and behave the same, knowing the vehicle they are gesturing to is an autonomous vehicle and not a human driver? And what gestures will still be relevant? To `Hail' an autonomous taxi, one may order only through their phone. For now, we assume pedestrians behave and use gestures towards the AV like a human driver.

\subsection{Gesture Annotation}
Annotating the videos could be done in many ways, especially regarding direction, which could also be challenging for the VLM or an LLM to understand. Directions could be explained from both the ego driver's and the pedestrian's perspective using the road.

As the captions are supposed to be combined as a prompt for an LLM to decide on an action, the captions should include enough information to assist with this decision. Assuming the caption is accurate, interpretation of the gesture should be enough for the LLM to avoid including a physical description of the pedestrian, which is less quantizable to discrete control decisions for the autonomous agent. However, if the model does not understand the complexity of the real-life scenario and misinterprets, the LLM can still interpret the gesture using the body analysis, depending on the accuracy of the VLM. Our study also highlights that sentence comparison is difficult to use as a proper evaluation, as it can be challenging to find a `language' that expresses all important properties of the body in the context to get the whole picture. Annotating the captions of the pedestrian can be tricky and must follow specific rules, as direction alone can have meaning from multiple perspectives, describing it as first- or third-person.
Additionally, `Left' and `Right' are not always enough to be precise, and can easily be misunderstood as another perspective or direction, and can be especially confounding in multiview input situations \cite{greer2023ensemble}. Using clock time or compass directions allows a broader scope of precision. This precision, however, may not be required and would be inefficient token utilization. In formal settings, directions do not require as much precision, whereas in informal settings, especially in emergencies, they require high precision.

We highlight some common characteristics here: P1 (`Forward', `Follow', `Hail') mentions the specific arm, uses ``me" as an ego driver, and does not always include interpretation. P2 (`Idle', `Left') changes from ``person", ``guy", and ``subject" when talking about the pedestrian, states perspective of left and right, (\textit{the driver's left side}). P3 (`Reverse', `Stop + pass') uses `person' as the pedestrian. P4 (`Stop, go') states both perspectives from the driver and the pedestrian. We highlight these differences to illustrate the inherent ambiguity in interpretation and language describing an instructive gesture.

\subsection{Embedded Similarity - Method 1}
The cosine similarity had some issues in terms of `Extended' information. Especially Qwen habitually expanded the caption with additional and sometimes hallucinated information. Indeed, it would not be directly similar to the ground truth, but as an LLM would interpret the correct gesture even with additional information, it should be evaluated as similar. However, this would not be possible if the caption contained too many tokens for the LLM to gather all the information, and it is also inefficient; the goal is a complete and compact representation of the relevant information.

Additionally, to validate the metric selection further, the number of captions could be extended to reduce noise even more. Adding one more caption to each level found a slight improvement towards the ideal metric. With more samples, utilizing, e.g., a boxplot instead of a barplot would enable more accurate statistical analysis. Specifically, `Extended' should be punished according to the length of the sentence.

\subsection{Classification - Method 2}
Requesting the VLM to respond with a single classification from possible classes may eventually be the most reliable way of evaluating task-specific capabilities. At least before continuing with more complex evaluation methods. However, our study shows that the VLMs are incapable of accurate classification. The models had a limited understanding, with output generalizing mainly to `Stop' gestures.

This evaluation is only 0.70 accurate even with human expert captions. This was possibly due to the low number of 8 frames, making it difficult to interpret. Also, some classes were similar, like `Forward' and `Follow', which can be challenging to distinguish.

\subsection{Reconstruction - Method 3}
Using MSE to compute the error for the movement has some limitations left to future research regarding hand gestures that can change the intent of a gesture significantly, especially in relation to arm position and with regard to the movement of various pose keypoints. This would be accommodated by weighting the smaller body parts, but an open question is with what ratios? Further, temporal alignment of most-similar poses is another important area for future development of such a reconstruction metric. Forcing both to stand on a specific mark would eliminate the distance and location variable, to focus more on the gesture itself.

This evaluation method computes the pedestrian's exact position, build, and movement, making an exact reenactment very difficult. Also, to a point where the details to reenact it correctly are not necessary to interpret the gesture. Reconstructing the human movements could also be done using Video Generative Models (VGMs), to enable efficient data generation. This would also show how well an LLM interprets information and avoids human unconscious or cultural interpretation.
\section{Concluding Remarks}
In conclusion, this study evaluated VLM's capability to recognize and caption human traffic gestures in the format of longer descriptions and interpretations. This was evaluated across three evaluation methods, each varying in abstraction, reasoning, and response details. Throughout the evaluation, the method results show that currently-trained VLMs are unreliable in capturing human traffic gestures with one individual participant in the frame.

The evaluation methods provide a range of evaluation foci. They can be applied in multiple evaluation studies containing information comparison in sentences, abstract human movement description forwarding, and concrete categorical human gesture classification. 

\subsection{Future Research}
Expanding this study would include varying static and temporal inputs, additional models, and videos with varying but realistic pedestrians and scenes. Separate studies would look into varying window sizes and frame rates, and how the VLMs would behave on only crops of the individual pedestrian. Alternative methods to evaluate captions should be explored, such as utilizing an LLM to compare generated captions with ground truth or identifying missing information within the captions as a basis for evaluation. 

Advancing VLMs in zero-shot to caption human movement and traffic gestures could be enforced by using pose models to help explain poses to the VLM or by augmenting the video by projecting poses upon the video.

Overall, this research highlights the relevance but difficulty of the task of gesture understanding for autonomous systems to safely navigate in driving environments where human interpretation and interaction are necessary. Future research in both model performance and evaluation can drive the development of interpretable and robust human-cooperative autonomous driving systems.
{
    \small
    \bibliographystyle{ieeetr}
    \bibliography{main}

\begin{thebibliography}{10}

\bibitem{salzmann2020trajectron++}
T.~Salzmann, B.~Ivanovic, P.~Chakravarty, and M.~Pavone, ``Trajectron++: Dynamically-feasible trajectory forecasting with heterogeneous data,'' in {\em Computer Vision--ECCV 2020: 16th European Conference, Glasgow, UK, August 23--28, 2020, Proceedings, Part XVIII 16}, pp.~683--700, Springer, 2020.

\bibitem{prakash2021multi}
A.~Prakash, K.~Chitta, and A.~Geiger, ``Multi-modal fusion transformer for end-to-end autonomous driving,'' in {\em Proceedings of the IEEE/CVF conference on computer vision and pattern recognition}, pp.~7077--7087, 2021.

\bibitem{greer2021trajectory}
R.~Greer, N.~Deo, and M.~Trivedi, ``Trajectory prediction in autonomous driving with a lane heading auxiliary loss,'' {\em IEEE Robotics and Automation Letters}, vol.~6, no.~3, pp.~4907--4914, 2021.

\bibitem{alofi2024pedestrian}
A.~Alofi, R.~Greer, A.~Gopalkrishnan, and M.~Trivedi, ``Pedestrian safety by intent prediction: A lightweight lstm-attention architecture and experimental evaluations with real-world datasets,'' in {\em 2024 IEEE Intelligent Vehicles Symposium (IV)}, pp.~77--84, IEEE, 2024.

\bibitem{roy2024doscenes}
P.~Roy, S.~Perisetla, S.~Shriram, H.~Krishnaswamy, A.~Keskar, and R.~Greer, ``doscenes: An autonomous driving dataset with natural language instruction for human interaction and vision-language navigation,'' {\em arXiv preprint arXiv:2412.05893}, 2024.

\bibitem{dey2017pedestrian}
D.~Dey and J.~Terken, ``Pedestrian interaction with vehicles: roles of explicit and implicit communication,'' in {\em Proceedings of the 9th international conference on automotive user interfaces and interactive vehicular applications}, pp.~109--113, 2017.

\bibitem{rasouli2019autonomous}
A.~Rasouli and J.~K. Tsotsos, ``Autonomous vehicles that interact with pedestrians: A survey of theory and practice,'' {\em IEEE transactions on intelligent transportation systems}, vol.~21, no.~3, pp.~900--918, 2019.

\bibitem{alshami2024coool}
A.~K. AlShami, A.~Kalita, R.~Rabinowitz, K.~Lam, R.~Bezbarua, T.~Boult, and J.~Kalita, ``Coool: Challenge of out-of-label a novel benchmark for autonomous driving,'' {\em arXiv preprint arXiv:2412.05462}, 2024.

\bibitem{atakishiyev2023explaining}
S.~Atakishiyev, M.~Salameh, H.~Babiker, and R.~Goebel, ``Explaining autonomous driving actions with visual question answering,'' in {\em 2023 IEEE 26th International Conference on Intelligent Transportation Systems (ITSC)}, pp.~1207--1214, IEEE, 2023.

\bibitem{keskar2025evaluating}
A.~Keskar, S.~Perisetla, and R.~Greer, ``Evaluating multimodal vision-language model prompting strategies for visual question answering in road scene understanding,'' in {\em Proceedings of the Winter Conference on Applications of Computer Vision}, pp.~1027--1036, 2025.

\bibitem{shao2024lmdrive}
H.~Shao, Y.~Hu, L.~Wang, G.~Song, S.~L. Waslander, Y.~Liu, and H.~Li, ``Lmdrive: Closed-loop end-to-end driving with large language models,'' in {\em Proceedings of the IEEE/CVF Conference on Computer Vision and Pattern Recognition}, pp.~15120--15130, 2024.

\bibitem{greer2024towards}
R.~Greer and M.~Trivedi, ``Towards explainable, safe autonomous driving with language embeddings for novelty identification and active learning: Framework and experimental analysis with real-world data sets,'' {\em arXiv preprint arXiv:2402.07320}, 2024.

\bibitem{greer2025language}
R.~Greer, B.~Antoniussen, A.~M{\o}gelmose, and M.~Trivedi, ``Language-driven active learning for diverse open-set 3d object detection,'' in {\em Proceedings of the Winter Conference on Applications of Computer Vision}, pp.~980--988, 2025.

\bibitem{xu2024drivegpt4}
Z.~Xu, Y.~Zhang, E.~Xie, Z.~Zhao, Y.~Guo, K.-Y.~K. Wong, Z.~Li, and H.~Zhao, ``Drivegpt4: Interpretable end-to-end autonomous driving via large language model,'' {\em IEEE Robotics and Automation Letters}, 2024.

\bibitem{chen2024end}
L.~Chen, P.~Wu, K.~Chitta, B.~Jaeger, A.~Geiger, and H.~Li, ``End-to-end autonomous driving: Challenges and frontiers,'' {\em IEEE Transactions on Pattern Analysis and Machine Intelligence}, 2024.

\bibitem{madan2024foundation}
N.~Madan, A.~M{\o}gelmose, R.~Modi, Y.~S. Rawat, and T.~B. Moeslund, ``Foundation models for video understanding: A survey,'' {\em Authorea Preprints}, 2024.

\bibitem{CaVLA}
H.~Arai, K.~Miwa, K.~Sasaki, Y.~Yamaguchi, K.~Watanabe, S.~Aoki, and I.~Yamamoto, ``Covla: Comprehensive vision-language-action dataset for autonomous driving,'' {\em arXiv preprint arXiv:2408.10845}, 2024.

\bibitem{greer2023robust}
R.~Greer, A.~Gopalkrishnan, J.~Landgren, L.~Rakla, A.~Gopalan, and M.~Trivedi, ``Robust traffic light detection using salience-sensitive loss: Computational framework and evaluations,'' in {\em 2023 IEEE Intelligent Vehicles Symposium (IV)}, pp.~1--7, IEEE, 2023.

\bibitem{greer2022salience}
R.~Greer, J.~Isa, N.~Deo, A.~Rangesh, and M.~M. Trivedi, ``On salience-sensitive sign classification in autonomous vehicle path planning: Experimental explorations with a novel dataset,'' in {\em Proceedings of the IEEE/CVF Winter Conference on Applications of Computer Vision}, pp.~636--644, 2022.

\bibitem{greer2023salient}
R.~Greer, A.~Gopalkrishnan, N.~Deo, A.~Rangesh, and M.~Trivedi, ``Salient sign detection in safe autonomous driving: Ai which reasons over full visual context,'' in {\em 27th International Technical Conference on the Enhanced Safety of Vehicles (ESV) National Highway Traffic Safety Administration}, no.~23-0333, 2023.

\bibitem{zhao2024drivellava}
R.~Zhao, Q.~Yuan, J.~Li, Y.~Fan, Y.~Li, and F.~Gao, ``Drivellava: Human-level behavior decisions via vision language model,'' {\em Sensors (Basel, Switzerland)}, vol.~24, no.~13, p.~4113, 2024.

\bibitem{amir2017low}
A.~Amir, B.~Taba, D.~Berg, T.~Melano, J.~McKinstry, C.~Di~Nolfo, T.~Nayak, A.~Andreopoulos, G.~Garreau, M.~Mendoza, {\em et~al.}, ``A low power, fully event-based gesture recognition system,'' in {\em Proceedings of the IEEE conference on computer vision and pattern recognition}, pp.~7243--7252, 2017.

\bibitem{escalera2013multi}
S.~Escalera, J.~Gonz{\`a}lez, X.~Bar{\'o}, M.~Reyes, O.~Lopes, I.~Guyon, V.~Athitsos, and H.~Escalante, ``Multi-modal gesture recognition challenge 2013: Dataset and results,'' in {\em Proceedings of the 15th ACM on International conference on multimodal interaction}, pp.~445--452, 2013.

\bibitem{multi-modal-gesture-recognition}
B.~Hamner, Isabelle, LoPoal, sescalera, and xbaro, ``Multi-modal gesture recognition.'' \url{https://kaggle.com/competitions/multi-modal-gesture-recognition}, 2013.
\newblock Kaggle.

\bibitem{ruffieux2014survey}
S.~Ruffieux, D.~Lalanne, E.~Mugellini, and O.~Abou~Khaled, ``A survey of datasets for human gesture recognition,'' in {\em Human-Computer Interaction. Advanced Interaction Modalities and Techniques: 16th International Conference, HCI International 2014, Heraklion, Crete, Greece, June 22-27, 2014, Proceedings, Part II 16}, pp.~337--348, Springer, 2014.

\bibitem{wiederertraffic}
J.~Wiederer, A.~Bouazizi, U.~Kressel, and V.~Belagiannis, ``Traffic control gesture recognition for autonomous vehicles. in 2020 ieee,'' in {\em RSJ International Conference on Intelligent Robots and Systems (IROS)}, pp.~10676--10683.

\bibitem{kobzarev2025gestllm}
O.~Kobzarev, A.~Lykov, and D.~Tsetserukou, ``Gestllm: Advanced hand gesture interpretation via large language models for human-robot interaction,'' {\em arXiv preprint arXiv:2501.07295}, 2025.

\bibitem{damonlpsg2024videollama2}
Z.~Cheng, S.~Leng, H.~Zhang, Y.~Xin, X.~Li, G.~Chen, Y.~Zhu, W.~Zhang, Z.~Luo, D.~Zhao, {\em et~al.}, ``Videollama 2: Advancing spatial-temporal modeling and audio understanding in video-llms,'' {\em arXiv preprint arXiv:2406.07476}, 2024.

\bibitem{damonlpsg2025videollama3}
B.~Zhang, K.~Li, Z.~Cheng, Z.~Hu, Y.~Yuan, G.~Chen, S.~Leng, Y.~Jiang, H.~Zhang, X.~Li, {\em et~al.}, ``Videollama 3: Frontier multimodal foundation models for image and video understanding,'' {\em arXiv preprint arXiv:2501.13106}, 2025.

\bibitem{Qwen2-VL}
P.~Wang, S.~Bai, S.~Tan, S.~Wang, Z.~Fan, J.~Bai, K.~Chen, X.~Liu, J.~Wang, W.~Ge, {\em et~al.}, ``Qwen2-vl: Enhancing vision-language model's perception of the world at any resolution,'' {\em arXiv preprint arXiv:2409.12191}, 2024.

\bibitem{dao2023flashattention2fasterattentionbetter}
T.~Dao, ``Flashattention-2: Faster attention with better parallelism and work partitioning,'' {\em arXiv preprint arXiv:2307.08691}, 2023.

\bibitem{reimers2019sentence}
N.~Reimers and I.~Gurevych, ``Sentence-bert: Sentence embeddings using siamese bert-networks,'' {\em arXiv preprint arXiv:1908.10084}, 2019.

\bibitem{salton1983introduction}
G.~Salton, ``Modern information retrieval,'' {\em (No Title)}, 1983.

\bibitem{zhang2019bertscore}
T.~Zhang, V.~Kishore, F.~Wu, K.~Q. Weinberger, and Y.~Artzi, ``Bertscore: Evaluating text generation with bert,'' {\em arXiv preprint arXiv:1904.09675}, 2019.

\bibitem{papineni2002bleu}
K.~Papineni, S.~Roukos, T.~Ward, and W.-J. Zhu, ``Bleu: a method for automatic evaluation of machine translation,'' in {\em Proceedings of the 40th annual meeting of the Association for Computational Linguistics}, pp.~311--318, 2002.

\bibitem{jaccard1901etude}
P.~Jaccard, ``{\'E}tude comparative de la distribution florale dans une portion des alpes et des jura,'' {\em Bull Soc Vaudoise Sci Nat}, vol.~37, pp.~547--579, 1901.

\bibitem{banerjee2005meteor}
S.~Banerjee and A.~Lavie, ``Meteor: An automatic metric for mt evaluation with improved correlation with human judgments,'' in {\em Proceedings of the acl workshop on intrinsic and extrinsic evaluation measures for machine translation and/or summarization}, pp.~65--72, 2005.

\bibitem{lin2004rouge}
L.~Chin-Yew, ``Rouge: A package for automatic evaluation of summaries,'' in {\em Proceedings of the Workshop on Text Summarization Branches Out, 2004}, 2004.

\bibitem{wolf-2020-transformers}
T.~Wolf, L.~Debut, V.~Sanh, J.~Chaumond, C.~Delangue, A.~Moi, P.~Cistac, T.~Rault, R.~Louf, M.~Funtowicz, {\em et~al.}, ``Transformers: State-of-the-art natural language processing,'' in {\em Proceedings of the 2020 conference on empirical methods in natural language processing: system demonstrations}, pp.~38--45, 2020.

\bibitem{greer2023ensemble}
R.~Greer and M.~Trivedi, ``Ensemble learning for fusion of multiview vision with occlusion and missing information: Framework and evaluations with real-world data and applications in driver hand activity recognition,'' {\em arXiv preprint arXiv:2301.12592}, 2023.

\end{thebibliography}
}


\end{document}